\pdfoutput=1

\documentclass[11pt]{article}

\usepackage{emnlp2021}
\usepackage{graphicx}
\usepackage{times}
\usepackage{latexsym}
\usepackage{hyperref}

\usepackage[T1]{fontenc}

\usepackage[utf8]{inputenc}

\usepackage{microtype}
\usepackage{graphicx}
\usepackage{titling}

\begin{document}
\title{GenPlot: Increasing the Scale and Diversity of Chart Derendering Data}
\author{Brendan Artley\\
  Independent Researcher \\
  Vancouver, BC, Canada \\
  \texttt{brendan.artley@gmail.com} \\
}
\date{\today}
\maketitle

\begin{abstract}
Vertical bars, horizontal bars, dot, scatter, and line plots provide a diverse set of visualizations to represent data. To understand these plots, one must be able to recognize textual components, locate data points in a plot, and process diverse visual contexts to extract information. In recent works such as Pix2Struct, Matcha, and Deplot, OCR-free chart-to-text translation has achieved state-of-the-art results on visual language tasks. These results outline the importance of chart-derendering as a pre-training objective, yet existing datasets provide a fixed set of training examples. In this paper, we propose GenPlot; a plot generator that can generate billions of additional plots for chart-derendering using synthetic data.
\end{abstract}

\section{Introduction}

Traditionally, OCR-aware methods such as LayoutLM \citep{LayoutLM}, PresSTU \citep{kil2022prestu}, PaLI \citep{chen2023pali}, and ChartBERT \citep{akhtar-etal-2023-reading} have been used to extract information from plots. While these models can accurately extract text, they require a dataset of labeled components which can be expensive to obtain. Additionally, plots do not always represent numerical components exactly (ie. Scientific Notation, Percent, etc.), and therefore post-processing is required to extract numerical information.

In recent works like Donut \citep{kim2022ocrfree}, Pix2Struct \citep{lee2022pix2struct}, Matcha \citep{liu2023matcha}, and Deplot \citep{liu2023deplot}, OCR-free chart-to-text translation methods are used. Donut focuses on document understanding, whereas pix2struct aims to provide a generic pre-trained checkpoint for many downstream visual language tasks. Matcha and Deplot are concurrent to Pix2Struct as the models use the same underlying architecture. The models all require comprehensive datasets for pre-training tasks to achieve state-of-the-art results on the PlotQA \citep{plotqa} and ChartQA \citep{chartqa} benchmarks. It is computationally expensive to obtain these large datasets for pre-training, and this is where GenPlot can help.

GenPlot provides a framework to generate billions of possible plot combinations for chart derendering tasks. GenPlot is a standalone Python script built using Matplotlib \citep{matplotlib} that can generate bar, scatter, line, and dot plots. The configuration of the hyperparameters in the script can be modified as each user sees fit. By default, we use similar categorical labels determined using GloVe \citep{glove} embeddings and randomly generated numerical data. We also provide a pre-generated set of 500,000 plots that can be used in place of the generator. GenPlot provides a way for researchers to quickly increase the scale of data for chart derendering tasks. 

\section{Related Work}

Recent works such as PlotQA \citep{plotqa}, ChartQA \citep{chartqa}, Matcha \citep{liu2023matcha}, and DVQA \citep{dvqa} generate plots with a focus on Visual Question Answering (VQA) \citep{agrawal2016vqa}. PlotQA provides bar, line, and scatter plots. CharQA provides bar, line, and pie plots, and DVQA provides bar plots. It is unclear which plot types are generated for Matcha pretraining, as the data is not available in the public domain. The result of these works is high-quality question-answer pairs, rather than a means to generate large-scale datasets.

An existing work for chart generation is FigureQA \citep{kahou2018figureqa}. This source provides synthetically generated data, bounding boxes, and question-answer pairs. FigureQA also provides 4 different plot types, and the authors have released the code for data generation. The generation process is limited to 100 colors, and fixed chart components such as gridlines, labels, and legends. 

GenPlot extends the work of FigureQA by providing a means to generate a larger and more diverse set of plots. It does this through random color sampling, and variability in margins, grids, ticks, labels, plot sizes, and more. To our knowledge, there is no existing framework to generate plots with such a high degree of variability, which motivated us to build this framework. In table~\ref{tab:DataStats}, we compare our dataset with existing works. Generate+ indicates whether the source provides the ability to generate new plots, "\#" and Tables is the number of plots in the dataset. For "\#" Plot Types we count the number of unique plot types (ie. Bar, Line, Plot, Pie, etc.). Orientation of the chart type is not considered. For example, horizontal bar plots and vertical bar plots are counted as one. 


\begin{table*}
\centering
\begin{tabular}{lcccccccc}
\hline
\textbf{Dataset} & \textbf{Generate+} & \textbf{\# Tables} & \textbf{Bar} & \textbf{Line} & \textbf{Pie} & \textbf{Scatter} & \textbf{Dot} & \textbf{Unique Plot Types} \\
\hline
ChartQA & No & 22k & 1 & 1 & 1 & 0 & 0 & 3 \\
PlotQA & No & 224k & 1 & 1 & 0 & 1 & 0 & 3 \\
MATCHA & No & 270k & 1 & 1 & 1 & 0 & 0 & 3 \\
DVQA & No & 300k & 1 & 0 & 0 & 0 & 0 & 1  \\
FigureQA & No & 100k & 1 & 1 & 1 & 1 & 0 & 5 \\
\hline
GenPlot (ours) & Yes & 500k & 1 & 1 & 0 & 1 & 1 & 5\\
\hline
\end{tabular}
\caption{Dataset Statistics}
\label{tab:DataStats}
\end{table*}

\section{Metadata and Generation}

Metadata for each plot is stored in a similar format to pix2struct chart-to-table models \citep{lee2022pix2struct}. For example, the string may look like “0 | 1 <0x0A> 1 | 2 <0x0A> 2 | 7”. In this string, there are 3 data points, (0,1), (1,2), and (2,7). The first value in each pair is the x value, and the second value is the y value. The x and y values are separated by the “|” character in the string. Each pair is then separated by the “<0x0A>” character sequence. All data points are stored in order from left to right, with the expectation of horizontal bar plots which are stored in a top-down fashion. Regardless of the data type on each axis, the format of the metadata stays the same.

\subsection{Text Generation}

Plots can contain text in the title, subtitles, and as categorical labels. To obtain groups of related words we utilized GloVe \citep{glove} embeddings. First, we set a predefined list of common objects like “toothbrush”, “coffee”, “notebook”, etc. Then, we sampled the 25 most similar words for each of the objects and added these to a vocabulary list. Any words that contained non-ASCII characters were not included. We performed this step twice which resulted in 42744 groups of similar words. We sampled in this way to ensure that label groups were related rather than randomly selected words. 

For each plot, the main title and x-axis title contains a sample of 3 to 7 words from a random group. The y-axis title contains a sample of 1-4 words. We set this range to reduce the chance of title and tick label overlap during the plot generation process.

For categorical labels, we use a list of place names, months, days, and part-numerical strings. The place names include a list of 2123 countries, regions, counties, and states. Months and days can be generated in numerous ways and occasionally contain numerical values. For example, the months and days could appear in the following formats: “Jan”, “December”, “Jan-Feb”, “Apr-10”,  “Tues”, “Friday”, etc. The part-numerical strings appear as follows: “10-20”, “30-40”, etc. We included part-numerical strings as categorical labels to create "difficult" examples.

\subsection{Numerical Generation}

The numerical data generation is done using a random polynomial sampler and a random linear sampler implemented in Numpy \citep{numpy}. Each sampler generates sequences of integers or floating point numbers and adds a small degree of Gaussian noise to the set of values. The values are then scaled by a factor between 0.01 and 1,000,000. Occasionally, random points are scaled again to simulate outliers in the data. Plot-specific modifications are also made to ensure the data was suitable for each plot type. 

\section{Plot Types and Styles}

In this section, we outline the default settings and parameter combinations in the plot generation process. Moreover, each plot is generated using a programming tool called Matplotlib \citep{matplotlib}. We select 8 unique styles, and 9 font families as default parameters. At the time of generation, the style and font family are randomly selected. We also randomly remove ticks, grids, and spines from each plot.

To generate the data point colors, we randomly sample RGB values between 40 and 200. This ensures that the data points can be seen on light and dark backgrounds, and gives more color variability to the plots. Finally, we enforce graph conventions from the Benetech - Making Graphs Accessible Competition \citep{benetech-making-graphs-accessible}. These conventions were in place so that generated plots are representative of non-generated plots.

\subsection{Bar plot}

There are 200,000 bar plots in the pre-generated data. Half of these are vertical bar plots, and the other half are horizontal bars. These plot types are very similar, with only the axes being flipped. We discuss the bar plots in the context of vertical bars for the remainder of this section. The number of generated bars is between 2 and 20, with oversampling to sizes of around 6 bars. Variation is added to the spacing between bars and the plot margins. The X-axis labels are integers or strings, and the Y-axis values are numerical. No value on the Y-axis can be 200 times greater than the minimum value in each set of values. Each X-axis label corresponds to a single bar in the plot.

\begin{figure}[ht]
\centering
\includegraphics[width=\columnwidth]{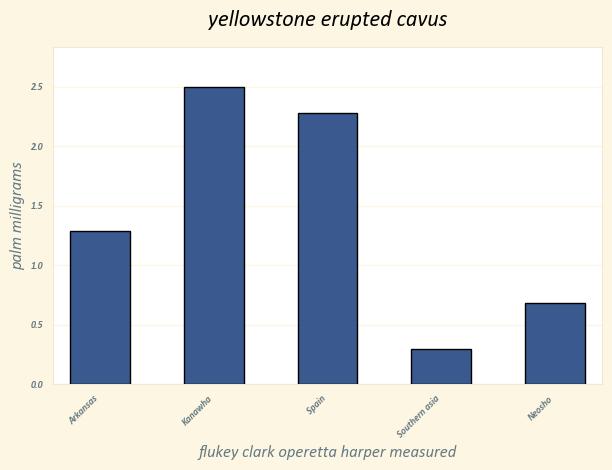}
\caption{Vertical Bar Plot}
\label{fig:VBar1}
\end{figure}

\begin{figure}[ht]
\centering
\includegraphics[width=\columnwidth]{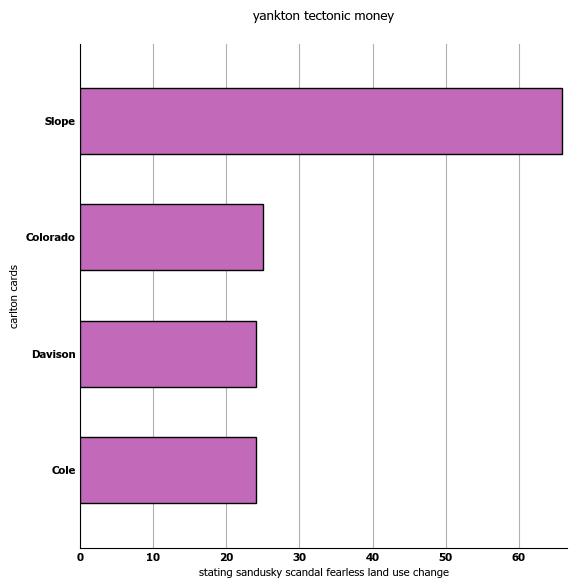}
\caption{Horizontal Bar Plot}
\label{fig:HBar1}
\end{figure}

\subsection{Scatter plot}

There are 100,000 scatter plots in the pre-generated data. The number of generated points in each plot is between 3 and 86. The two main variations of the scatter plot are randomly sampled points, and points that follow a path-like pattern. For a high number of randomly sampled points, values likely overlap. Therefore we implement a custom sampler to ensure that any two points do not completely overlap. The line-like scatter plot is generated using the polynomial sampler and provides a version of the plot type similar to the line plot. The X-axis labels and the Y-axis labels are numerical.

\begin{figure}[ht]
\centering
\includegraphics[width=\columnwidth]{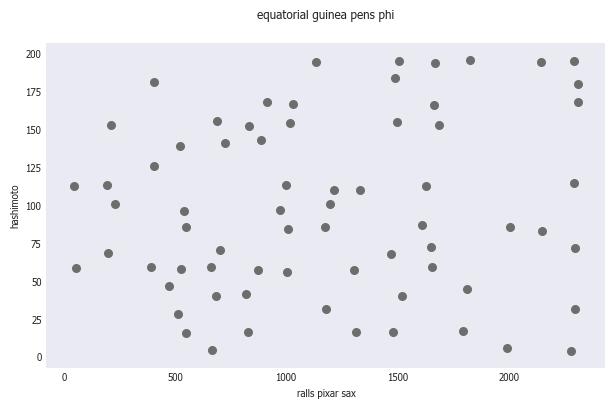}
\caption{Scatter Plot}
\label{fig:Scatter1}
\end{figure}

\subsection{Line plot}

There are 100,000 line plots in the pre-generated data. The number of generated points is between 2 and 20, with oversampling to sizes of around 7 values. Values found on the X-axis are either dates or integer values in ascending order. The Y-axis is strictly numerical. One unique characteristic of the line plot is that X-axis labels do not always correspond to data points. This is done to ensure that OCR-free systems do not just read the data labels, and instead learn from the entire context. Additionally, dots are randomly added to the line, and random smoothing is occasionally applied. This is done to increase variability.

\begin{figure}[ht]
\centering
\includegraphics[width=\columnwidth]{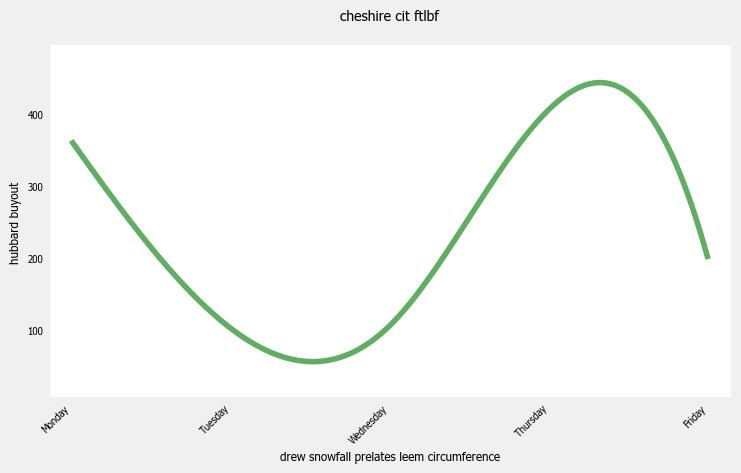}
\caption{Line Plot}
\label{fig:Line1}
\end{figure}

\subsection{Dot Plot}

\begin{figure}[ht]
\centering
\includegraphics[width=\columnwidth]{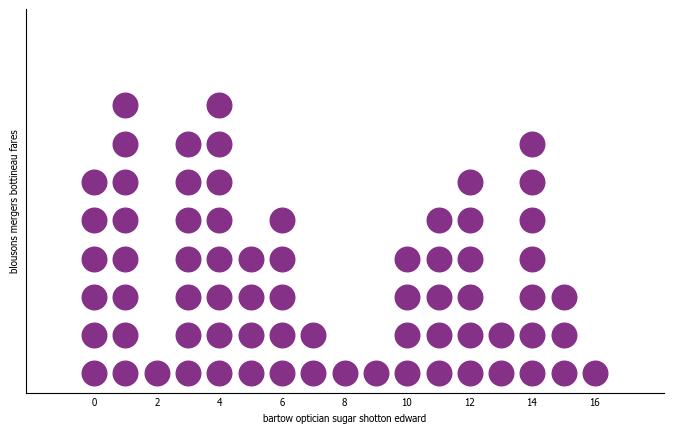}
\caption{Dot Plot}
\label{fig:DotPlot1}
\end{figure}

There are 100,000 dot plots in the pre-generated data. The number of generated points is between 2 and 21 values. Values on the Y-axis are always integers between 1 and 10, and the Y-axis values can be strings or numerical. A unique characteristic of the dot plots is that numerical X-axis labels are not always present, as they can be inferred from surrounding labels. Additionally, Y-axis ticks are sometimes completely removed, as counts can be inferred from the plot. See table~\ref{fig:DotPlot1} for an example of this. Also, as of this writing, there is no standardized Matplotlib function for dot plots, so we implement a custom dot plot function using the scatter function.

\section{Conclusion}

We have proposed a framework for large-scale plot generation for chart-derendering. We use a variety of coloring, spacing, and sampling options to yield a diverse generation for 4 plot types. We provide a pre-generated set of 500,000 plots which exceeds the size of existing datasets. The code for GenPlot can be found on GitHub (link), along with instructions on how to get started. The dataset containing the pre-generated plots can be found on Kaggle.
\\ \\
Links: \href{https://github.com/brendanartley/GenPlot}{Source Code}, \href{https://www.kaggle.com/datasets/brendanartley/genplot}{Pre-Generated Data}

\subsection{Limitations}

Though we built a diverse plot generator, there are still many plot types that we did not implement. For example, we did not include pie plots, histograms, or treemaps. It remains up for debate whether adding more plot types would be beneficial for chart-derendering as a pre-training task. Additionally, we added safeguards to reduce the likelihood of label overlap, but there is still a possibility of this happening in newly generated plots. We were unable to find a way to validate this without a human in the loop. 

\subsection{Ethics Statement}

During the data generation process, we considered several ethical issues. Firstly, we acknowledge Gender and Race bias found in GloVe embeddings. Occupations and degrees of similarity between words are stereotyped. We do not condone the stereotypes found in this model, and recognize how this is a result of its training data.

\label{sec:bibtex}

\bibliography{custom}
\bibliographystyle{acl_natbib}

\appendix



\end{document}